\documentclass{midl} 

\usepackage{array}
\usepackage{booktabs}
\usepackage{multirow}
\usepackage{adjustbox}
\usepackage{xcolor}

\jmlrvolume{-- 374}
\jmlryear{2026}
\jmlrworkshop{Full Paper -- MIDL 2026 submission}
\editors{Accepted for publication at MIDL 2026}

\title[BrainATCL]{BrainATCL: Adaptive Temporal Brain Connectivity Learning for Functional Link
Prediction and Age Estimation}

\midlauthor{\Name{Yiran Huang\nametag{$^{1}$}} 
\Email{yh87@njit.edu}\\
\Name{Amirhossein Nouranizadeh\nametag{$^{1}$}} \Email{an857@njit.edu}\\
\Name{Christine Ahrends\nametag{$^{2}$}} \Email{christine.ahrends@ndcn.ox.ac.uk}\\
\Name{Mengjia Xu\midljointauthortext{Corresponding Author}\nametag{$^{1}$}} \Email{mx6@njit.edu}\\
\addr $^{1}$ Department of Data Science, New Jersey Institute of Technology, Newark, NJ, USA \\
\addr $^{2}$ Nuffield Department of Clinical Neurosciences, University of Oxford, UK
}

\begin{document}

\maketitle

\begin{abstract}
Functional Magnetic Resonance Imaging (fMRI) is an imaging technique widely used to study human brain activity. fMRI signals in areas across the brain transiently synchronise and desynchronise their activity in a highly structured manner, even when an individual is at rest. These functional connectivity dynamics may be related to behaviour and neuropsychiatric disease. To model these dynamics, temporal brain connectivity representations are essential, as they reflect evolving interactions between brain regions and provide insight into transient neural states and network reconfigurations. However, conventional graph neural networks (GNNs) often struggle to capture long-range temporal dependencies in dynamic fMRI data. To address this challenge, 
we propose {\it BrainATCL}, an unsupervised, nonparametric framework for adaptive temporal brain connectivity learning, enabling functional link prediction and age estimation. Our method dynamically adjusts the lookback window for each snapshot based on the rate of newly added edges. Graph sequences are subsequently encoded using a \texttt{GINE-Mamba2} backbone to learn spatial-temporal representations of dynamic functional connectivity in resting-state fMRI data of 1,000 participants from the Human Connectome Project. To further improve spatial modeling, we incorporate brain structure and function-informed edge attributes, i.e., the left/right hemispheric identity and subnetwork membership of brain regions, enabling the model to capture biologically meaningful topological patterns. We evaluate our BrainATCL on two tasks: functional link prediction and age estimation. The experimental results demonstrate superior performance and strong generalization, including in cross-session prediction scenarios.
\end{abstract}


\section{Introduction}

Functional magnetic resonance imaging (fMRI) measures brain activity non-invasively via blood-oxygen-level-dependent (BOLD) signals, serving as a key neuroimaging technique for studying neural processes and supporting disease diagnosis~\cite{perovnik2023functional,biswal2025history}. In resting-state fMRI, where no explicit task is performed, spontaneous synchrony and desynchrony across brain regions reveal structured, large-scale networks associated with behavior, cognition, aging, and neurological or psychiatric conditions~\cite{biswal1995functional, shen2015resting, smith2013functional}. Functional brain connectivity can be modeled from fMRI signals as either static functional connectivity (sFC) or dynamic functional connectivity (dFC)~\cite{
power2011functional}. sFC assumes constant interactions between brain regions throughout the fMRI scan, and is usually measured by the Pearson or partial correlation. Significant works utilize sFC to study brain abnormalities linked to various brain diseases, e.g., Alzheimer's disease~\cite{xu2020new,xu2021graph, baker2024hyperbolic} or Autism Spectrum Disorder~\cite{zhang2022classification}. However, as human brain transitions through dynamic functional states, dFC has emerged as a powerful alternative to sFC by capturing time-varying interactions between brain regions, offering insights into transient neural states and network reconfigurations~\cite{ahrends2025dynamic}. 

Capturing temporal dynamics in dFCs remains challenging due to variability from sampling errors, physiological artifacts, and arousal fluctuations, amplified by the low signal-to-noise ratio inherent in fMRI~\cite{laumann2024challenges}. Recent advances in dFC modeling primarily fall into three categories: {\bf 1) Mode/Component decomposition-based methods}, which model dFC by decomposing signals or discrete-time connectivity patterns into interpretable bases, using techniques like temporal covariance EVD (eigenvalue decomposition)~\cite{alteriis2025dysco},  principle component analysis (PCA)~\cite{leonardi2013principal}, Fisher kernel~\cite{ahrends2025predicting}, and the Koopman operator~\cite{turja2023deepgraphdmd}. These methods often assume linearity and may struggle to generalize across datasets. {\bf 2) Clustering-based methods}, which segment time-varying FC matrices into discrete brain states using methods like k-means~\cite{van2024dynamic} or deep clustering~\cite{spencer2022using}. However, these methods struggle to capture smooth or overlapping transitions between dynamic connectivity states and reply on ad-hoc parameters.
{\bf 3) Graph-based dynamic modeling}, which integrate static graph neural networks (GNNs) with different temporal modules, such as temporal convolutional networks (TCNs)~\cite{gadgil2020spatio, azevedo2022deep}, recurrent neural networks (RNNs)~\cite{cao2022modeling,s25010156}, attention-based transformers~\cite{kim2021learning, yu2024long, wang2024stnagnn}, to capture complex spatiotemporal dynamics of dFCs. 

Despite the strong expressive power, existing methods mostly rely on fixed context windows, suffer from high computational overhead, and overlook neuroscience-informed node and edge attributes -- limiting generalizability across subjects or sessions. They also struggle to capture long-range temporal dependencies, hindering accurate and scalable modeling of spatio-temporal patterns in dFCs. However, the recent introduction of Mamba~\cite{dao2024transformers}, a state-space model (SSM) with linear complexity, presents a promising direction for efficient and scalable sequence modeling. More related works on SSM can be found from section~\ref{sec:related-works}.

In this work, to integrate neuroscience-inspired priors, long-range temporal modeling, and adaptive mechanisms to improve the accuracy and robustness of dFC analysis, we propose {\bf BrainATCL}, an unsupervised, nonparametric framework for adaptive temporal brain connectivity representation learning from resting-state fMRI data. It consists of four key innovations: (1) Adaptive determination of the temporal lookback window at each time point based on the rate of newly formed links, enabling the model to capture meaningful transitions in functional connectivity. (2) Incorporation of structure- and function-informed edge attributes, i.e., hemispheric identity and subnetwork membership, to enhance spatial representations and capture biologically meaningful topological patterns. (3) Encoding of adaptively generated graph sequences via the \texttt{GINE-Mamba2} backbone, which integrates edge-aware spatial message passing with efficient long-range temporal modeling to learn expressive temporal graph embeddings. (4) Evaluation of the proposed BrainATCL framework on two different tasks, \textit{functional link prediction (in both within and cross-session settings)} and {\it age estimation}, using the Human Connectome Project (HCP) dataset including 1,000 subjects. Experimental results demonstrate its effectiveness and strong generalizability in modeling dFCs, outperforming both static and dynamic graph learning baselines.

\section{Related Work}
\label{sec:related-works}

Graph neural networks (GNNs) have emerged as a powerful tool for both spatial and temporal graph modeling via incorporating conventional sequence models such as recurrent neural networks~\cite{xu2022dyng2g}, long-short term memory (LSTM)~\cite{s25010156}, and \cite{WANG2025130582} further incorporate a diffusion connection strategy and adaptive self-attention fusion to enhance the GNN-LSTM model stability and spatiotemporal representation. Recently, transformers have shown great potential for temporal graph modeling via the self-attention mechanism~\cite{varghese2024transformerg2g}.
\cite{yu2024long} develop a brain graph transformer to capture long-range dependencies within brain networks. \cite{wang2024stnagnn} incorporated the node-level attention algorithm for information aggregation on ROI-based brain graphs. \cite{kan2022brainnetworktransformer} propose a brain network transformer that models brain graphs with fixed node order and connection profiles and introduce an orthonormal clustering readout. However, attention-based approaches often suffer from high computational cost with quadratic complexity.

State space models (SSMs)~\cite{gu2021efficiently}, particularly advanced Mamba model~\cite{ dao2024transformers}, have gained increasing interest for their linear complexity and efficiency compared to transformers~\cite{vaswani2017attention}. Several studies have attempt to apply Mamba for dFC modeling, e.g., Brain-GM~\cite{wang2024learning} marks the first integration of the state space model into dynamic brain graph representation learning; FST-Mamba~\cite{wei2025hierarchical} is designed for high-dimensional dFC features that can learn dynamic connectivity patterns and predict biological and cognitive outcomes. \cite{zrimek2025dynstgmambadynamicspatiotemporalgraph}'s method learns dynamic spatial connections, adjusting the graph structure based on temporal variations in movement.  
However, most existing methods rely on fixed temporal windows and overlook neuroscience-informed edge priors, motivating our adaptive framework development with dynamic context selection and biologically grounded edge features


\section{Methodology}
\subsection{Data Preparation} 
{\bf fMRI preprocessing:} we used resting-state fMRI recordings of 1,001 healthy subjects (aged 22-35) from the Young Adult S1200 release of the Human Connectome Project 
~\cite{van2013wu,smith2013resting}. Briefly, data were acquired on a 3T MRI scanner. For each participant, resting-state functional scans were acquired in four sessions of 14 mins 33 sec duration each, using multiband echo planar imaging (EPI) at an acceleration factor of 8, a repetition time (TR) of 0.72 sec, and a spatial resolution of $2\times2\times2$mm. The data were preprocessed as described in \cite{smith2013resting} using minimal spatial preprocessing followed by ICA-based temporal preprocessing and high-pass filtering using a 2000 sec cut-off. Since the voxel-level timecourses are too high-dimensional, we parcellated the brain using the Schaefer parcellation~\cite{schaefer2018local} with 100 regions of interest (ROIs). We chose this relatively low-dimensional parcellation, since estimating dFC from more fine-grained parcellations can lead to overfitting~\cite{ahrends2022data}. Timecourses were extracted as the first principal component within each ROI. Timecourses from eight ROIs were removed as noise $z\text{-score} > 6$, resulting in 92 ROIs. The input to our experiment is the BOLD time series of $N=92$ brain regions for $P=1000$ subjects over $T=1200$ time steps, and each subject has $Q=4$ imaging sessions. Hence, we denote the entire time series by $\mathcal{X} \in \mathbb{R}^{N\times P \times Q \times T}$. 
All detailed symbols are provided in Appendix Table~\ref{tab:symbols}.

\noindent
{\bf Sliding window-based dFC construction:} we compute the Pearson correlation coefficient between pairs of brain regions with a sliding window of length $w$ and stride $s$. Each FC matrix is preprocessed by zeroing the diagonal and setting negative correlations to zero, followed by thresholding off-diagonal entries with $\tau$ to retain only positive connections and capture temporal dFC patterns. The choice of threshold $\tau$ is essential: larger $\tau$ yields more dynamic graphs with higher edge turnover, while a smaller $\tau$ produces smoother graphs that may retain more noise. The impact of different $\tau$ values on model performance is discussed in Section~\ref{sec:experiments}. To further understand how $\tau$ affects the temporal dynamics of graphs, we provide an in-depth analysis in
Appendix~\ref{appendix:tau}.

The resulting data is a set of all discrete-time dynamic graphs for each subject $p$ and session $q$ denoted by $\mathcal{G}=\{G^{p,q}|p\in \{0, \dots, 1000\}, q\in \{0, \dots, 3\} \}$, where $G^{p, q}=(G^{p, q}_0, \dots, G^{p, q}_k, \dots G^{p, q}_{K-1})$, $k$ is the sequence index and $K = \lfloor \frac{1200 - w}{s} \rfloor + 1$. Note that each functional connectivity graph is represented by its corresponding adjacency, node and edge feature matrices coupled with their target variables $\mathbf{y}^p$, $G^{p, q}_k=(\mathbf{X}_k^{p, q}, \mathbf{A}_k^{p, q}, \mathbf{E}^{p, q}_k, \mathbf{y}^p)$, where $\mathbf{X}_k^{p, q} \in \mathbb{R}^{N \times F_V}$, $\mathbf{E}^{p, q}_k \in \mathbb{R}^{M \times F_E}$ and $\mathbf{A}_k^{p, q} \in \{0, 1\}^{N \times N}$, and $F_V, F_E$ are the number of input node and edge features, respectively.
\begin{figure}[htbp]
  \centering
  \includegraphics[width=.9\textwidth]{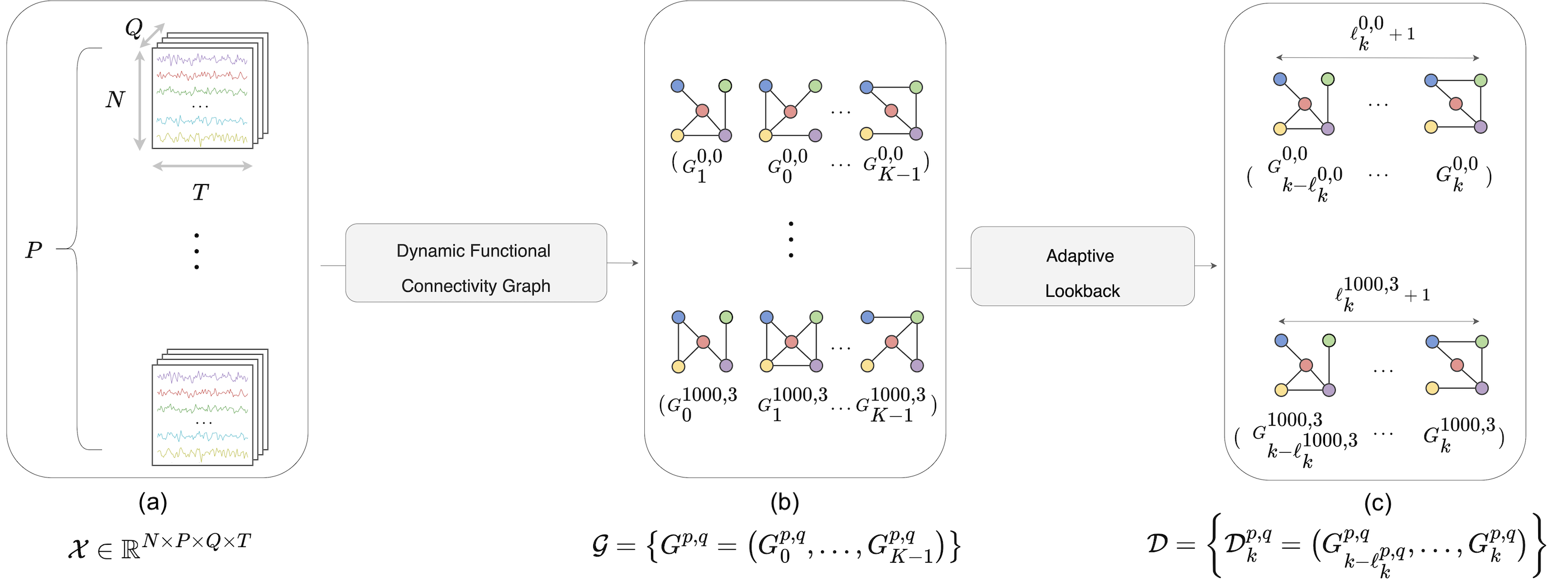}
  \caption{Pipeline for dFC construction and adaptive graph sequence generation. \textbf{(a)} The original 4D fMRI BOLD signals $\mathcal{X}\in \mathbb{R}^{N \times P \times Q \times T}$ (with $N$, $P$, $Q$, $T$ denoting ROIs, subjects, imaging sessions, and time points). \textbf{(b)} We construct dFC $G^{p,q}$ as a graph sequence of length $K$ for subject $p$ and session $q$, all subjects' dFC set is denoted as $\mathcal{G}$. \textbf{(c)} For training the models, we designed adaptive lookback strategy as a function of the temporal novelty index (Eq.~\ref{eq:novelty}) as a heuristic to determine temporal context $\mathcal{D}^{p,q}_k$ dynamically for each graph snapshot at time step $k$,  see detailed computation of adaptive lookback $\ell^{p, q}_k$ in Section~\ref{sec:adaptive_lookback}.}
  \label{fig:data_preprocessing}
\vspace{-.25in}
\end{figure}

\subsection{Problem Definition}
The goal is to learn dFC representations that effectively and efficiently encode their spatiotemporal information. 
Mathematically, the goal is to learn an embedding function $f$ that maps an arbitrary graph sequence $G^{p,q} \in \mathcal{G}$ into an embedding matrix $\mathbf{H}^{p, q} = f(G^{p,q}) \in \mathbb{R}^{N \times D}$, where each row corresponds to a node embedding vector of dimension $D$. The learned embeddings $\mathbf{H}^{p,q}$ support two downstream tasks: (1) {\it dynamic link prediction}, where the next adjacency matrix is estimated as
$\hat{\mathbf{A}}^{p,q}_K = g(\mathbf{H}^{p,q})$ with $g: \mathbb{R}^{N\times D} \rightarrow \{0,1\}^{N\times N}$ as the link prediction function, and $\hat{\mathbf{A}}_K^{p, q}$ is the estimated adjacency matrix of the graph at time step $K$; (2) {\it Age prediction}, a dynamic graph regression task, the subject’s age is estimated as $\hat{y}^p=r(\mathbf{H}^{p, q})$, where $r: \mathbb{R}^{N \times D} \rightarrow \mathbb{R}_+$ is the regression function. 

\vspace{-.13in}

\subsection{The Proposed BrainATCL Framework}
\label{sec:BrainATCL-Framework}

\begin{figure}[ht]
  \centering
  \includegraphics[width=0.9\textwidth]{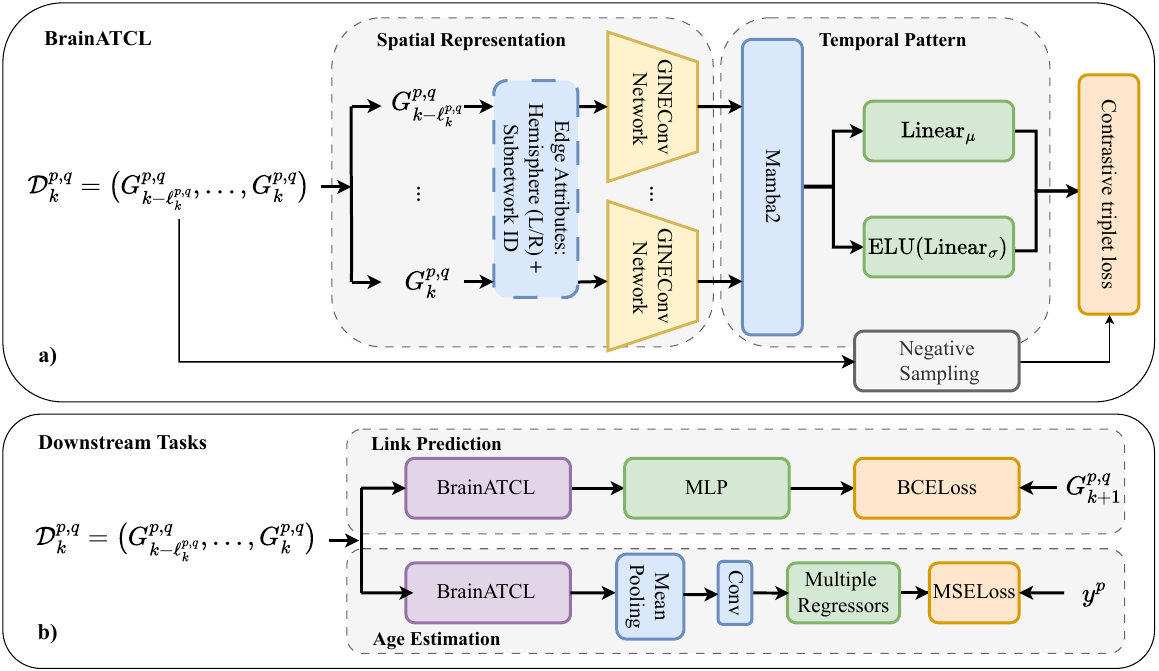}
  \caption{Illustration of our proposed BrainATCL framework for adaptive temporal brain connectivity embedding and two downstream tasks.
  \textbf{(a)} The BrainATCL is trained by minimizing the contrastive triplet loss. It combines a GINEConv-based GNN to encode structural information with edge attributes, followed by a Mamba-2 model for temporal aggregation across the graph sequence. The output of Mamba-2 is fed to linear layers to represent each node as a multivariate Gaussian distribution. \textbf{(b)} After pre-training the BrainATCL, the input graph sequence is processed and its node embeddings are used for link prediction classifier and age prediction regressor. These downstream task models are trained separately using binary cross-entropy (BCE) and mean-squared error (MSE) losses.}
  \label{fig:models}
\vspace{-.25in}
\end{figure}

In our proposed BrainATCL framework, we approach dFC representation learning and downstream machine learning tasks in two phases followed by data preprocessing illustrated in Fig.~\ref{fig:data_preprocessing}.
The workflow of our BrainATCL is shown in Fig.~\ref{fig:models}.

\subsubsection{Novelty-based adaptive lookback computation.}\label{sec:adaptive_lookback}
One of the main components of our method is the adaptive computation of the temporal context for each static graph within the dynamic functional connectivity.
We define the temporal context as the set of historical graphs prior to each static graph within the graph sequence.
We characterize the temporal context of each static graph $G^{p,q}_k$ using the adaptive lookback number $\ell^{p,q}_k$, which indicates the number of past relevant graphs for the static graph corresponding to subject $p$ in session $q$ at time step $k$.
We are motivated to compute the temporal context for each static graph during data preprocessing to introduce an inductive bias. Specifically, our framework explicitly models temporal graph dynamics by aggregating historical context of adaptive length, which alleviates the need for the model to implicitly learn long-range temporal dependencies.

Our intuition is based on the \textit{novelty index} of the dynamic graph, defined as the average ratio of new edges at each time step:
\begin{equation}
n_k = \frac{1}{K} \sum_{k=1}^{K} \frac{\left| E_k \setminus E_k^{\text{seen}} \right|}{\left| E_k \right|},
\label{eq:novelty}
\end{equation}
where $E_k$ represents the set of edges at time step $k$, and $E_k^{\mathrm{seen}}$ represents the set of edges observed in previous time steps $1,\dots,k-1$.

We aim to compute the lookback number $\ell^{p,q}_k$ as a function of the novelty index such that the dynamic graph is assigned a shorter lookback at time steps with rapid novel changes and a longer lookback at time steps where the graph evolves more slowly.
This intuition is supported by the observation that a dynamic graph undergoing frequent novel changes behaves similarly to a sequence of random graphs, making historical context less relevant, whereas slow variations in the graph indicate that its structure depends on a longer historical context.
We employ the novelty index of the dynamic graph to quantify structural variations and compute the lookback number as a function of this index.

Specifically, let $(n^{p,q}_0,\dots,n^{p,q}_{K-1})$ be the sequence of novelty indices for the dynamic graph $G^{p,q}=(G^{p,q}_0,\dots,G^{p,q}_{K-1})$, where each novelty index is computed as defined in Eq.~\ref{eq:novelty}.
We begin with a desired range of lookback values, $[\ell_{\min}, \ell_{\max}]$, so that $B = \ell_{\max} - \ell_{\min} + 1$ is the number of discrete lookback values.
Given the sequence of novelty indices, we compute the $(B+1)$-quantiles as:
\begin{equation}
    \alpha^{p, q}_b = \mathrm{Quantile}\bigl(n^{p,q}_0,\dots,n^{p,q}_{K-1};\,\tfrac{b}{B}\bigr),\enspace b=\{0,\dots,B\},
\end{equation}
so $\alpha^{p,q}_0$ and $\alpha^{p,q}_B$ are equal to the minimum and maximum novelty indices in the sequence, respectively.
To compute the adaptive lookback for each time step $k$ of the dynamic graph $G^{p,q}$, we first find the unique bin index $b \in {0,\dots,B-1}$ to which the corresponding novelty index belongs: $\alpha^{p, q}_b \leq n^{p, q}_k < \alpha^{p, q}_{b+1},$
and assign the adaptive lookback value $\ell^{p, q}_k = \ell_{\max} - b$.
Thus, graphs with the lowest novelty index ($b=0$) receive the longest lookback value $\ell^{p,q}_k = \ell_{\max}$, while those with the highest novelty index ($b=B-1$) receive the shortest lookback value $\ell^{p,q}_k = \ell_{\min}$. We determine an adaptive lookback number for each dynamic graph, resulting in the final dataset. $\mathcal{D}$ structured as follows:
\begin{equation}
    \mathcal{D} = \left\{  \mathcal{D}^{p, q}_k=\bigl(G^{p,q}_{k-\ell^{p,q}_k}, \dots, G^{p,q}_{k}\bigr)  \;\middle|\;  \begin{aligned}    & \ell_{\max} \le k \le K\\    & p \in \{0, \dots, 1000\}\\    & q \in \{0, \dots, 3\}  \end{aligned}\right\}
\end{equation}
\subsubsection{Encoding of dFCs via GINE-Mamba} In the first step of modeling, we employ a contrastive learning approach to encode the spatial and temporal information contained in the dFC graph into the embedding vector space.
Inspired by previous work \cite{bojchevski2017deep, parmanand2024comparative}, we model each node in the graph sequence as a multivariate Gaussian distribution by parameterizing its mean vector and covariance matrix.
Specifically, for each static graph at time step $k$ in the graph sequence, we encode its structural information using a graph neural network composed of $L$ $\mathtt{GINEConv}$ \cite{hu2019strategies} layers:
\begin{equation}
\begin{aligned}
  \mathbf{z}_{k, i}^{\{p, q\}, l}
    &= h_{\theta_l}\Bigl(
         (1+\epsilon)\,\mathbf{z}_{k, i}^{\{p, q\}, l-1}  
    + \sum_{j \in \mathcal{N}^{p, q}_k(i)} \mathrm{ReLU}\bigl(\mathbf{z}_{k, j}^{\{p, q\}, l-1} + \mathbf{e}^{p, q}_{k,ji}\bigr)
      \Bigr),
\end{aligned}
\end{equation}
where $p$ and $q$ are indicators of the subject and session, respectively. $\mathbf{z}_{k, i}^{\{p, q\}, l} \in \mathbb{R}^{D_l}$ is the $l^\text{th}$ layer’s structural embedding of node $i$ at time step $k$, $\epsilon$ is a learnable parameter, $l = 1, \dots, L$ is the number of $\mathtt{GINEConv}$ layers, $\mathcal{N}^{p,q}_k(i)$ is the set of node $i$'s neighbors in the graph sequence at time step $k$, and $\mathbf{e}^{p,q}_{k,ji}$ is the edge feature vector of edge $ji$ at time step $k$. Note that the input to the first layer is the node feature vector, i.e., $\mathbf{z}_{k,i}^{{p,q},0} = \mathbf{x}^{p,q}_{k,i}$. To incorporate biologically meaningful spatial priors into the model, we encode structure- and function-informed edge attributes by leveraging the hemispheric identity and subnetwork membership of the nodes connected by each edge. Specifically, for every edge $ji$, we obtain the hemisphere and subnetwork labels of both endpoint nodes, embed these categorical labels into learnable vectors, and concatenate them to form the final edge attribute.

After encoding the structural information of static graphs within the graph sequence $\mathcal{D}^{p,q}_k = \bigl(G^{p,q}_{k-\ell^{p,q}_k}, \dots, G^{p,q}_{k}\bigr)$, we obtain a sequence of embeddings for each node in the graph, i.e., $(\mathbf{z}_{k-\ell^{p,q},i}^{{p,q},L}, \dots, \mathbf{z}_{k,i}^{{p,q},L})$ for time steps $k - \ell^{p,q}_k, \dots, k$. 
Inspired by recent advances in state-space models for sequence modeling \cite{gu2023mamba, dao2024transformers}, we employ the $\mathtt{Mamba2}$ model to encode the temporal evolution of structural embeddings into spatiotemporal embeddings.
\begin{equation}
    \mathbf{h}^{p, q}_{k, i}= \mathtt{Mamba2}(\mathbf{z}_{k-\ell^{p, q}_k, i}^{\{p, q\}, L}, \dots, \mathbf{z}_{k, i}^{\{p, q\}, L}).
\end{equation}

The spatiotemporal embeddings $\mathbf{h}^{p,q}{k,i} \in \mathbb{R}^{D\text{emb}}$ are then fed into two separate linear layers to compute the mean and covariance of each node:
\begin{equation}
    \begin{aligned}
        \mathbf{\mu}^{p, q}_{k, i} &= \mathtt{Linear}_\mu(\mathbf{h}^{p, q}_{k, i}), \\
        \mathbf{\sigma}^{p, q}_{k, i} &= \text{ELU}\bigl(\mathtt{Linear}_\sigma(\mathbf{h}^{p, q}_{k, i})\bigr)+1+\sigma.
    \end{aligned}
\end{equation}

The mean and variance vectors $\boldsymbol{\mu}^{p,q}_{k,i}$ and $\boldsymbol{\sigma}^{p,q}_{k,i}$ are in $\mathbb{R}^{D_\text{emb}}$, characterizing the Gaussian distribution of node $i$'s embedding in $\mathcal{D}^{p,q}_k$ as $\mathcal{N}(\boldsymbol{\mu}^{p,q}_{k,i}, \text{diag}(\boldsymbol{\sigma}^{p,q}_{k,i}))$. We add 1 to the \text{ELU} output to ensure covariance positivity; $\sigma$ is a small constant ($10^{-14}$) for numerical stability.

We train the BrainATCL by optimizing a contrastive triplet loss.
For each reference node $i$ of the static graph $G^{p,q}_k$ in the graph sequence $\mathcal{D}^{p,q}_k$, we extract its direct neighbors as positive samples $i^{\text{near}}$ and the nodes two hops away from the reference node as negative samples $i^{\text{far}}$.
Reference nodes and their corresponding positive and negative samples form a set of triplets $\mathcal{T}^{p,q}_k = {(i, i^{\text{near}}, i^{\text{far}}) \mid i \in G^{p,q}_k}$ used in the contrastive representation learning.

The contrastive loss function aims to minimize the distance between the embeddings of adjacent nodes while maximizing the distance between the embeddings of non-adjacent nodes.
We define the contrastive loss for the graph sequence sample $\mathcal{D}^{p, q}_k$ as
\begin{equation}
    \mathcal{L}_{\mathcal{D}^{p, q}_k}=\frac{1}{\ell^{p, q}_k + 1}\sum_{G^{p, q}_k \in \mathcal{D}^{p, q}_k} \sum_{ \mathcal{T}^{p, q}_k}[E^2{(i, i^{\text{near}})}+e^{-E{(i, i^{\text{far}})}}],
\end{equation}
where the energy function $E$ computes the Kullback-Leibler divergence between the corresponding Gaussian distribution representations of the input nodes, i.e.,
\begin{equation}
    E(i, j)=\text{KL}(\mathcal{N}(\mu_i, \Sigma_i)||\mathcal{N}(\mu_j, \Sigma_j)).    
\end{equation}
The loss objective of the BrainATCL is the sum of embedding losses over all graph sequence samples: $\mathcal{L} = \sum_{\mathcal{D}}\mathcal{L}_{\mathcal{D}^{p, q}_k}$.

\subsubsection{Evaluation on downstream prediction tasks.}
Once the pre-trained BrainATCL is optimized using contrastive learning, we embed the dFC graph and use the embeddings for downstream tasks:

For the link prediction task, we represent each edge as the concatenation of the mean vectors of its corresponding endpoint node embeddings and feed this representation to a multilayer perceptron (MLP), i.e., $\hat{\mathbf{A}}_{k+1, ij}^{p, q} = \mathtt{MLP}([\mu^{p, q}_{k, i}||\mu^{p, q}_{k, j}])$. 
Here, $\hat{\mathbf{A}}_{k+1, ij}^{p, q}$ represents the probability of observing the edge $(i, j)$ at time step $k+1$ for subject $p$ and session $q$, given the sequence of graphs $(G^{p,q}_{k-\ell^{p,q}_k}, \dots, G^{p,q}_{k}\bigr)$.
The $\mathtt{MLP}$ is trained using binary cross-entropy between predictions and the ground-truth edges at the next time step.

For age prediction, we use the mean of each node’s Gaussian distribution as its embedding and aggregate them into subject-level temporal representations using $\mathtt{MeanPooling}$ and a $\mathtt{Conv}$ module, then pass it through a $\mathtt{Regressor}$ layer to compute the predicted age, i.e., $\hat{y}^p=\mathtt{Regressor}(\mathtt{Conv}(\mathtt{MeanPooling}(\{ \mu^{p, q}_{k, i} \}_{i \in G^{p, q}_k})))$. (see Appendix~\ref{appendix:details} for details)

\vspace{-.16in}
\section{Experimental Results}
\subsection{Dataset}
We evaluated our approach on the HCP young adult dataset for two tasks: temporal link prediction and age prediction. In the link prediction task, we select the first 300 subjects to train the BrainATCL and conduct the link prediction task. The trained model is then used to generate dynamic functional connectivity (dFC) graph embeddings for a separate set of 300 subjects (not used in training the BrainATCL), which are subsequently used for the age prediction task.
\vspace{-.15in}

\subsection{Implementation Details}
We train BrainATCL under two settings: {\bf 1) Within-session:} each of the four sessions per subject is split by timestamps: the first 70\% of each session is used for training, the next 10\% for validation, and the remaining 20\% for testing. The training segments from all sessions and all subjects are concatenated to form the final training set; validation and test segments are concatenated similarly.
{\bf 2) Cross-session:} the first two sessions are used for training, the third for validation, and the fourth for testing. Given the temporal gap between the four sessions, we exclude the first $\ell_{\max}$ graphs at the beginning of each session when computing the adaptive lookback, preventing the model from referencing graphs across session boundaries and ensuring semantic consistency. See Appendix~\ref{appendix:details} for more implementation details.

\subsection{Evaluation Tasks and Metrics}
We assess the performance of BrainATCL on temporal link prediction task using five standard metrics -- mean average precision (MAP), mean reciprocal rank (MRR), area under the ROC curve (AUC), area under the precision--recall curve (PR-AUC), and F1-score. See Appendix~\ref{appendix:lp_metrics} for detailed definitions.
For age estimation task, the evaluation metrics are mean absolute error (MAE) and mean square error (MSE), both measured in years. We adopted five regression models for age prediction: {\it ElasticNet, Ridge, Lasso, Random Forest, and Support Vector Regression (SVR)}. For each model, we perform grid search to identify the optimal hyperparameters. The model with the best performance is selected based on validation results.

\subsection{Results}
\label{sec:experiments}

We conduct an ablation study under two training setups, systematically varying key hyperparameters such as lookback window, threshold, embedding size, and stride length. As shown in Appendix~\ref{appendix:Ablation_p} Fig.~\ref{fig:thred_embsize_stepsize}, both the mean and variance of MAP and MRR scores are reported for each configuration. 
Based on this comparison, we select the configuration with threshold 0.5, embedding size 128, adaptive lookback, and stride size 10 for the final model comparison. This setup provides a strong balance between within-session and cross-session performance, yielding high MAP and MRR scores across both evaluation scenarios. For the choice of temporal window size, \cite{allen2014tracking, hutchison2013resting} recommend windows of approximately $60s$ to ensure reliable estimation of time-varying correlations. Accordingly, we use a window size of $w=100$  $(72 s)$, which provides a principled balance between estimation reliability and sensitivity to transient resting-state fluctuations.

We also evaluate the impact of incorporating edge features into the BrainATCL model. We compare two variants: with and without edge features, and summarize the results for both within-session and cross-session evaluations in Table~\ref{tab:edge_ablation}.
Including edge features consistently improves performance across all five evaluation metrics, indicating that added relational information helps capture fine-grained variations in dynamic connectivity patterns.

\begin{table}[ht]
\centering
\vspace{-.15in}
\caption{Ablation study on the effect of edge features (including network and hemisphere labels of the two endpoint nodes) for temporal link prediction under within-session and cross-session settings.}
\label{tab:edge_ablation}

\resizebox{\textwidth}{!}{
\begin{tabular}{lccccc}
\toprule
\multicolumn{6}{c}{\textbf{Within-session}} \\
\midrule
\textbf{Model Variant}
& MAP$\uparrow$ & MRR$\uparrow$ & AUC$\uparrow$ & PR-AUC$\uparrow$ & F1$\uparrow$ \\
\midrule

w/o Edge Features 
& 0.7313 $\pm$ 0.0024 
& 0.6061 $\pm$ 0.0015 
& 0.9728 $\pm$ 0.0004 
& 0.7143 $\pm$ 0.0016 
& 0.6830 $\pm$ 0.0011\\

w/ Edge Features 
& \textbf{0.7681 $\pm$ 0.0012} 
& \textbf{0.6265 $\pm$ 0.0009} 
& \textbf{0.9746 $\pm$ 0.0001} 
& \textbf{0.7468 $\pm$ 0.0013} 
& \textbf{0.7050 $\pm$ 0.0008}\\

\bottomrule
\end{tabular}
}

\vspace{2mm}

\resizebox{\textwidth}{!}{
\begin{tabular}{lccccc}
\toprule
\multicolumn{6}{c}{\textbf{Cross-session}} \\
\midrule
\textbf{Model Variant}
& MAP$\uparrow$ & MRR$\uparrow$ & AUC$\uparrow$ & PR-AUC$\uparrow$ & F1$\uparrow$ \\
\midrule

w/o Edge Features 
& 0.7257 $\pm$ 0.0016 
& 0.6051 $\pm$ 0.0009 
& 0.9705 $\pm$ 0.0004 
& 0.7061 $\pm$ 0.0008 
& 0.6815 $\pm$ 0.0013 \\

w/ Edge Features 
& \textbf{0.7667 $\pm$ 0.0004} 
& \textbf{0.6194 $\pm$ 0.0006} 
& \textbf{0.9741 $\pm$ 0.0001} 
& \textbf{0.7352 $\pm$ 0.0016} 
& \textbf{0.6923 $\pm$ 0.0017}\\

\bottomrule
\end{tabular}
}

\end{table}


We further conducted an ablation study to evaluate the impact of different training objectives. In particular, we replace the contrastive loss with a VGAE-style reconstruction objective using inner-product decoding and KL regularization ~\cite{kipf2016variational}.
As shown in Table \ref{tab:cons_ablation}, the model trained with the variational loss consistently underperforms the proposed contrastive learning framework across all evaluation metrics. This indicates that purely generative reconstruction objectives are insufficient to capture discriminative temporal patterns in dynamic brain networks.

\begin{table}[ht]
\centering
\vspace{-.15in}
\caption{Ablation study on training objectives by replacing the proposed contrastive loss with a VGAE-style reconstruction objective for temporal link prediction under the within-session setting.}
\label{tab:cons_ablation}

\resizebox{\textwidth}{!}{
\begin{tabular}{lccccc}
\toprule
\multicolumn{6}{c}{\textbf{Within-session}} \\
\midrule
\textbf{Model Variant}
& MAP$\uparrow$ & MRR$\uparrow$ & AUC$\uparrow$ & PR-AUC$\uparrow$ & F1$\uparrow$ \\
\midrule

VGAE Reconstruction
& 0.7054 $\pm$ 0.0019 
& 0.5982 $\pm$ 0.0021 
& 0.9701 $\pm$ 0.0004 
& 0.6752 $\pm$ 0.0012 
& 0.6488 $\pm$ 0.0011 \\

Contrastive (BrainATCL)
& \textbf{0.7681 $\pm$ 0.0012} 
& \textbf{0.6265 $\pm$ 0.0009} 
& \textbf{0.9746 $\pm$ 0.0001} 
& \textbf{0.7468 $\pm$ 0.0013} 
& \textbf{0.7050 $\pm$ 0.0008}\\

\bottomrule
\end{tabular}
}

\end{table}

Under this optimal configuration, we evaluate BrainATCL against a comprehensive set of baselines (classical and recent graph models) on temporal link and age prediction. "Within-session" and "cross-session" refer to the two training setups. As shown in Table~\ref{tab:model_comparison}, BrainATCL achieves the best performance across all evaluation metrics, with MAP of 0.7681, MRR of 0.6265, AUC of 0.9746, PR-AUC of 0.7468, and F1-score of 0.7050, significantly outperforming all other baselines. The baselines include general graph neural networks - 
GCN~\cite{kipf2016semi} GAT~\cite{hamilton2017inductive}, GraphSAGE~\cite{velivckovic2017graph}, GIN~\cite{xu2018powerful}), GNN-RNNs variants (GCN-GRU, GAT-GRU, GIN-GRU, GraphSAGE-GRU)~\cite{seo2018structured,chen2022gc}, a dynamic graph variational method (VGRNN)~\cite{hajiramezanali2019variational}, and two state-of-the-art temporal graph models: TransformerG2G~\cite{varghese2024transformerg2g} and $\mathcal{GDG}$-Mamba~\cite{pandey2024comparative}. Notably, while $\mathcal{GDG}$-Mamba performs competitively, BrainATCL achieves better performance through the integration of biologically meaningful edge attributes and a novelty-aware adaptive lookback mechanism.

\begin{table}[htbp]
\centering
\setlength{\tabcolsep}{1.1mm} 
\caption{Benchmarking BrainATCL against multiple baselines for temporal link and age prediction under within-session and cross-session settings. 
MAP, MRR, AUCROC, PR-AUC, and F1 are used for link prediction, while MAE and MSE are used for age prediction. 
Reported values are means with standard deviations shown in parentheses. 
All comparisons are conducted under the same experimental configuration (stride=10, window=100, embedding=128, threshold=0.5).}

\resizebox{\textwidth}{!}{
\begin{tabular}{c|ccccc|cc}
\toprule
\multicolumn{8}{c}{\textbf{Within-session}} \\
\midrule
\textbf{Model}
& MAP$\uparrow$ & MRR$\uparrow$ & AUC$\uparrow$ & PR-AUC$\uparrow$ & F1$\uparrow$
& MAE$\downarrow$ & MSE$\downarrow$ \\
\midrule

GCN & 0.2838 {\scriptsize (0.0018)} & 0.2226 {\scriptsize (0.0009)} & 0.9101 {\scriptsize(0.0000)} & 0.2679 {\scriptsize(0.0001)} & 0.3557 {\scriptsize(0.0001)}
    & 3.09 {\scriptsize (0.16)} & 13.77 {\scriptsize (1.89)} \\

GraphSAGE & 0.3245 {\scriptsize (0.0234)} & 0.2576 {\scriptsize (0.0534)} & 0.9279 {\scriptsize(0.0000)} & 0.3017 {\scriptsize(0.0001)} & 0.4054 {\scriptsize(0.0002)}
    & 3.29 {\scriptsize (0.58)} & 16.58 {\scriptsize (5.42)} \\

GAT & 0.3601 {\scriptsize (0.0027)} & 0.2045 {\scriptsize (0.0038)} & 0.9112 {\scriptsize(0.0000)} & 0.3408 {\scriptsize(0.0001)} & 0.4233 {\scriptsize(0.0002)}
    & 3.16 {\scriptsize (0.49)} & 15.78 {\scriptsize (2.45)} \\

GIN & 0.3564 {\scriptsize (0.0124)} & 0.2308 {\scriptsize (0.0210)} & 0.9178 {\scriptsize(0.0000)} & 0.3326 {\scriptsize(0.0001)} & 0.4150 {\scriptsize(0.0000)}
    & 3.15 {\scriptsize (0.07)} & 14.70 {\scriptsize (1.23)} \\

GCN-GRU & 0.2991 {\scriptsize (0.0055)} & 0.2351 {\scriptsize (0.0037)} & 0.9172 {\scriptsize(0.0000)} & 0.2785 {\scriptsize(0.0000)} & 0.3622 {\scriptsize(0.0000)}
    & 3.08 {\scriptsize (0.54)} & 14.68 {\scriptsize (0.77)} \\

GraphSAGE-GRU & 0.4161 {\scriptsize (0.0041)} & 0.3752 {\scriptsize (0.0061)} & 0.9484 {\scriptsize(0.0000)} & 0.4043 {\scriptsize(0.0001)} & 0.4961 {\scriptsize(0.0000)}
    & 3.25 {\scriptsize (0.41)} & 16.85 {\scriptsize (3.93)} \\

GAT-GRU & 0.3765 {\scriptsize (0.0023)} & 0.2256 {\scriptsize (0.0167)} & 0.9183 {\scriptsize(0.0000)} & 0.3489 {\scriptsize(0.0001)} & 0.4176 {\scriptsize(0.0000)}
    & 3.06 {\scriptsize (0.06)} & 13.49 {\scriptsize (1.23)} \\

GIN-GRU & 0.3727 {\scriptsize (0.0061)} & 0.2446 {\scriptsize (0.0036)} & 0.9304 {\scriptsize(0.0000)} & 0.3356 {\scriptsize(0.0001)} & 0.4363 {\scriptsize(0.0000)}
    & 2.92 {\scriptsize (0.37)} & \textbf{10.77} {\scriptsize (2.15)} \\

VGRNN & 0.3987 {\scriptsize (0.0047)} & 0.2523 {\scriptsize (0.0021)} & 0.9112 {\scriptsize(0.0000)} & 0.3742 {\scriptsize(0.0001)} & 0.4399 {\scriptsize(0.0001)}
    & 3.32 {\scriptsize (0.36)} & 17.01 {\scriptsize (0.59)} \\

\midrule

TransformerG2G 
& 0.6615 {\scriptsize (0.0063)} 
& 0.5675 {\scriptsize (0.0036)} 
& 0.9730 {\scriptsize (0.0002)} 
& 0.6509 {\scriptsize (0.0006)} 
& 0.6786 {\scriptsize (0.0014)}
& 2.87 {\scriptsize (0.41)} & 12.18 {\scriptsize (2.86)} \\

$\mathcal{GDG}$-Mamba
& 0.7201 {\scriptsize (0.0034)} 
& 0.5951 {\scriptsize (0.0018)} 
& 0.9711 {\scriptsize (0.0001)} 
& 0.6978 {\scriptsize (0.0007)} 
& 0.6786 {\scriptsize (0.0012)}
& 2.84 {\scriptsize (0.48)} & 12.09 {\scriptsize (2.65)} \\

\textbf{BrainATCL}
& \textbf{0.7681} {\scriptsize (0.0012)} & \textbf{0.6265} {\scriptsize (0.0009)}
& \textbf{0.9746} {\scriptsize (0.0001)} & \textbf{0.7468} {\scriptsize (0.0013)} & \textbf{0.7050} {\scriptsize (0.0008)}
& \textbf{2.81} {\scriptsize (0.37)} & 11.73 {\scriptsize (2.57)} \\

\bottomrule
\end{tabular}
}

\vspace{2mm}

\resizebox{\textwidth}{!}{
\begin{tabular}{c|ccccc|cc}
\toprule
\multicolumn{8}{c}{\textbf{Cross-session}} \\
\midrule
\textbf{Model}
& MAP$\uparrow$ & MRR$\uparrow$ & AUC$\uparrow$ & PR-AUC$\uparrow$ & F1$\uparrow$
& MAE$\downarrow$ & MSE$\downarrow$ \\
\midrule

GCN & 0.2767 {\scriptsize (0.0002)} & 0.2262 {\scriptsize (0.0005)} & 0.9113 {\scriptsize (0.0000)} 
& 0.2652 {\scriptsize (0.0000)} 
& 0.3528 {\scriptsize (0.0000)}
    & 3.46 {\scriptsize (0.01)} & 16.38 {\scriptsize (0.79)} \\

GraphSAGE & 0.3127 {\scriptsize (0.0473)} & 0.2892 {\scriptsize (0.0674)} & 0.9166 {\scriptsize (0.0000)} 
& 0.2536 {\scriptsize (0.0001)} 
& 0.3581 {\scriptsize (0.0001)}
    & 3.64 {\scriptsize (0.59)} & 17.17 {\scriptsize (4.22)} \\

GAT & 0.3512 {\scriptsize (0.0018)} & 0.1825 {\scriptsize (0.0007)} & 0.9120 {\scriptsize (0.0000)} 
& 0.3393 {\scriptsize (0.0001)} 
& 0.4206 {\scriptsize (0.0000)}
    & 3.27 {\scriptsize (0.001)} & 15.79 {\scriptsize (0.25)} \\

GIN & 0.3423 {\scriptsize (0.0067)} & 0.2217 {\scriptsize (0.1298)} & 0.9215 {\scriptsize (0.0000)} 
& 0.3612 {\scriptsize (0.0001)} 
& 0.4370 {\scriptsize (0.0000)}
    & 3.04 {\scriptsize (0.74)} & 14.65 {\scriptsize (3.75)} \\

GCN-GRU & 0.2801 {\scriptsize (0.0035)} & 0.2285 {\scriptsize (0.0016)} & 0.9186 {\scriptsize (0.0000)} 
& 0.2778 {\scriptsize (0.0000)} 
& 0.3613 {\scriptsize (0.0000)}
    & 3.41 {\scriptsize (0.21)} & 15.53 {\scriptsize (0.67)} \\

GraphSAGE-GRU & 0.4173 {\scriptsize (0.0020)} & 0.3798 {\scriptsize (0.0054)} & 0.9496 {\scriptsize (0.0000)} 
& 0.4055 {\scriptsize (0.0001)} 
& 0.4974 {\scriptsize (0.0001)}
    & 3.44 {\scriptsize (0.31)} & 17.64 {\scriptsize (4.41)} \\

GAT-GRU & 0.3689 {\scriptsize (0.0025)} & 0.2058 {\scriptsize (0.0314)} & 0.9203 {\scriptsize (0.0000)} 
& 0.3507 {\scriptsize (0.0001)} 
& 0.4190 {\scriptsize (0.0000)}
    & 3.28 {\scriptsize (0.22)} & 15.98 {\scriptsize (2.01)} \\

GIN-GRU & 0.3813 {\scriptsize (0.0059)} & 0.2328 {\scriptsize (0.0097)} & 0.9321 {\scriptsize (0.0000)} 
& 0.3618 {\scriptsize (0.0000)} 
& 0.4372 {\scriptsize (0.0000)}
    & 2.97 {\scriptsize (0.52)} & \textbf{11.62} {\scriptsize (2.89)} \\

VGRNN & 0.3838 {\scriptsize (0.0034)} & 0.2489 {\scriptsize (0.0078)} & 0.9105 {\scriptsize (0.0000)} 
& 0.3755 {\scriptsize (0.0000)} 
& 0.4423 {\scriptsize (0.0000)}
    & 3.58 {\scriptsize (0.11)} & 17.39 {\scriptsize (0.76)} \\

\midrule

TransformerG2G
& 0.6643 {\scriptsize (0.0050)} & 0.5689 {\scriptsize (0.0031)} 
& 0.9714 {\scriptsize (0.0000)} 
& 0.6589 {\scriptsize (0.0010)} 
& 0.6799 {\scriptsize (0.0007)}
& 3.02 {\scriptsize (0.45)} & 13.44 {\scriptsize (3.09)} \\

$\mathcal{GDG}$-Mamba
& 0.7002 {\scriptsize (0.0029)} & 0.5898 {\scriptsize (0.0013)} 
& 0.9674 {\scriptsize (0.0001)} 
& 0.6700 {\scriptsize (0.0016)} 
& 0.6603 {\scriptsize (0.0008)}
& 2.83 {\scriptsize (0.38)} & 12.04 {\scriptsize (3.04)} \\

\textbf{BrainATCL}
& \textbf{0.7667} {\scriptsize (0.0004)} & \textbf{0.6194} {\scriptsize (0.0003)}
& \textbf{0.9741} {\scriptsize (0.0001)} & \textbf{0.7352} {\scriptsize (0.0016)} & \textbf{0.6923} {\scriptsize (0.0017)}
& \textbf{2.80} {\scriptsize (0.38)} & 11.68 {\scriptsize (2.97)} \\

\bottomrule
\end{tabular}
}

\label{tab:model_comparison}
\end{table}

Regarding explainability, since our cohort consists of healthy young adults with a relatively narrow age range (22–37 years), ROI- and edge-level attributions are expected to exhibit a low signal-to-noise ratio and limited translational relevance. To provide a preliminary analysis, we selected the two subjects with the largest age difference (22 and 37 years) and computed the cosine distance between their learned ROI embeddings. The resulting cross-subject differences were further grouped according to major functional networks, as shown in Appendix~\ref{appendix:explain} Fig.~\ref{fig:explain}. As illustrated in the figure, inter-subject variability is not uniformly distributed across networks, but is mainly concentrated in the visual, dorsal attention, and control systems, whereas somatomotor, limbic, and default-mode networks exhibit substantially smaller differences. These observations are consistent with prior resting-state functional connectivity studies reporting network-specific effects of aging, with pronounced changes in task-positive and visual networks~\cite{mowinckel2012network}. For more details, see Appendix~\ref{appendix:explain}.
Taken together, our node-level representation analysis suggests that cross-subject differences captured by BrainATCL are structured and network-dependent, reflecting biologically plausible patterns of age-related functional reorganization, even within a relatively narrow age range.

\section{Conclusion}
Dynamic functional connectivity analysis faces challenges from fixed context windows and the limited integration of neuroscience-informed priors. To address these issues, we propose \textbf{BrainATCL}, an unsupervised and adaptive framework that dynamically adjusts temporal context based on temporal novelty index and incorporates biologically meaningful edge attributes to enhance spatial modeling. Experiments on HCP data demonstrate its strong performance and generalizability across both link prediction and age estimation tasks. One limitation of the current framework is the lack of end-to-end training between the embedding model and downstream tasks. In future work, we plan to enhance the adaptivity of the framework by aligning temporal context selection more directly with brain activity dynamics, and to explore the identification of individual-specific brain modes during the resting state.

\clearpage  
\midlacknowledgments{This work is supported in part by DOE SEA-CROGS project (DESC0023191), AFOSR project(FA9550-24-1-0231), and the Grace Hopper AI Research Institute (GHAIRI) Seed Grant at NJIT. We also thank the computing resources provided by the High Performance Computing (HPC) facility at NJIT. Christine Ahrends is supported by the Carlsberg Foundation (CF23-1716). We would like to acknowledge Usama Pervaiz for his contribution to data curation. Data were provided by the Human Connectome Project, WU-Minn Consortium (Principal Investigators: David Van Essen and Kamil Ugurbil; 1U54MH091657) funded by the 16 NIH Institutes and Centers that support the NIH Blueprint for Neuroscience Research; and by the McDonnell Center for Systems Neuroscience at Washington University.}

\bibliography{midl26_374}

\newpage
\appendix

\begin{table}[h]
  \footnotesize                  
  \renewcommand{\arraystretch}{1.2}
  \caption{List of Symbols and Their Descriptions}
  \begin{adjustbox}{max width=\linewidth}
    \begin{tabular}{>{\bfseries}l p{0.65\linewidth}}
      \hline
      \textbf{Symbol} & \textbf{Description} \\
      \hline
      $N$                 & Number of nodes \\
      $M$                 & Number of edges \\
      $P$                 & Number of subjects \\
      $Q$                 & Number of sessions \\
      $T$                 & Number of time steps in BOLD time series \\
      $\mathcal{X}$       & BOLD time series \\
      $w$                 & Window size for computing the Pearson correlation coefficient \\
      $s$                 & Stride size for computing the Pearson correlation coefficient \\
      $\mathcal{G}$       & Set of all dynamic functional connectivity graphs \\
      $G^{p, q}$          & Dynamic functional connectivity graph for subject $p$ and session $q$ \\
      $K$                 & Number of static graphs in the dynamic functional connectivity \\
      $G^{p, q}_k$ & Static graph for subject $p$ and session $q$ at time step $k$\\
      $\mathbf{X}^{p, q}_k$ & Node feature matrix for subject $p$ and session $q$ at time step $k$ \\
      $\mathbf{E}^{p, q}_k$ & Edge feature matrix for subject $p$ and session $q$ at time step $k$ \\
      $\mathbf{A}^{p, q}_k$ & Adjacency matrix for subject $p$ and session $q$ at time step $k$ \\
      $\mathbf{y}^p$, $y^p$   & Target variable vector and scalar for subject $p$ \\
      $F_V$               & Number of node features \\
      $F_E$               & Number of edge features \\
      $f$                 & Embedding function \\
      $\mathbf{H}^{p, q}$ & Node embedding matrix for subject $p$ and session $q$ \\
      $\hat{\mathbf{A}}^{p, q}_K$ & Predicted adjacency matrix for subject $p$ and session $q$ at time step $K$ \\
      $g$                 & Link prediction function \\
      $r$                 & Age prediction function \\
      $\mathbf{z}_{k, i}^{\{p, q\}, l}$ & $l^{\text{th}}$ layer's spatial embedding vector of node $i$ for subject $p$, session $q$ and time index $k$\\
      $\mathcal{N}^{p, q}_k(i)$ & Set of neighbors of node $i$ for subject $p$ and session $q$ at time step $k$\\
      $\mathbf{e}^{p, q}_{k, ji}$ & Feature vector of edge $(j, i)$ for subject $p$ and session $q$ at time step $k$\\
      $L$ & Number of $\mathtt{GINEConv}$ layers\\
      $\mathbf{h}^{p, q}_{k, i}$ & Hidden state of Mamba2 for node $i$ at time step $k$ corresponding to subject $p$ and session $q$\\
      $\mathbf{\mu}^{p, q}_{k, i}, \mathbf{\sigma}^{p, q}_{k, i}$ & Mean and standard deviation of Gaussian distribution corresponding to the embeddings of node $i$ at time step $k$ for subject $p$ and session $q$\\
      $D_\text{emb}$ & Embedding dimension size\\
      $E(i, j) $ & Energy function that computes KL divergence between embeddings of nodes $i$ and $j$\\
      $ \mathcal{L}$ & Loss function for BrainATCL's training\\
      $\mathcal{L}_{\mathcal{D}^{p, q}_k}$ & Embedding loss corresponding to the sample $\mathcal{D}^{p, q}_k$\\      
      $n^{p, q}_k$ & Novelty index of the graph $G^{p, q}_k$\\
      $B$ & Number of discrete lookback values and number of quantile bins\\
      $\alpha^{p, q}_b$ & The $b^{\text{th}}$ quantile of novelty index for subject $p$ and session $q$\\
      $\ell^{p, q}_k$ & Adaptive lookback number for the graph $G^{p, q}_k$\\
      $ \mathcal{D}^{p, q}_k$ & Graph sequence sample with adaptive lookback corresponding to subject $p$, session $q$ and time step $k$\\
      \hline
    \end{tabular}
  \end{adjustbox}
  \label{tab:symbols}
\end{table}

\section{Impact of Threshold $\tau$ on Temporal Graph Dynamics}
\label{appendix:tau}
\begin{figure}[htbp]
  \centering
  \includegraphics[width=1\textwidth]{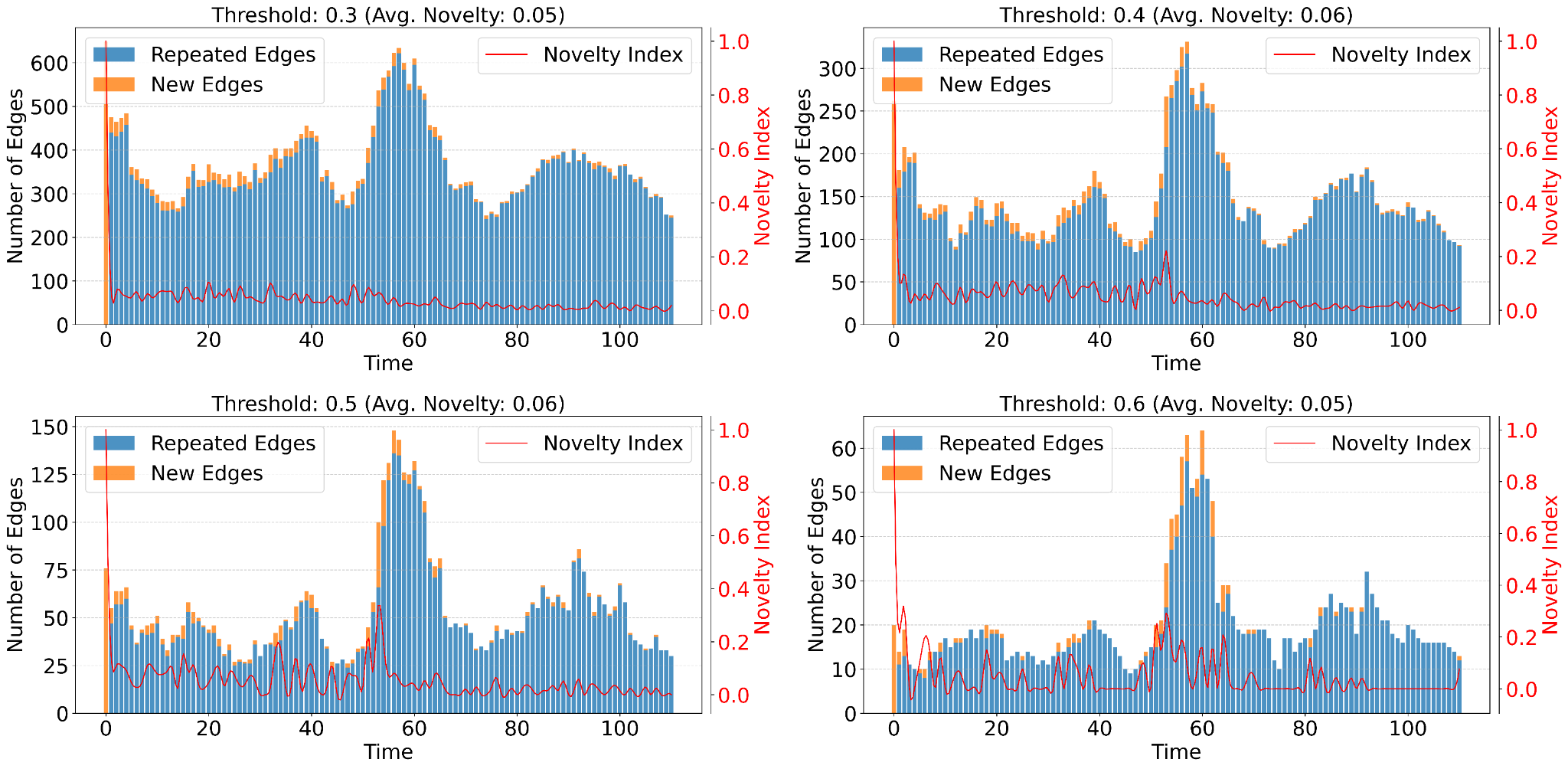}
  \caption{Temporal Edge Appearance (TEA) plots for subject 10, session 1 under different threshold values. For each time point, the blue bars indicate the number of repeated edges (i.e., edges that have appeared in any previous time point), while the orange bars represent newly appeared edges. The red line shows the novelty index at each time point, quantifying the proportion of new edges relative to total edges. The average novelty across the time series is reported in the title of each subplot.}
  \label{fig:TEA}
\end{figure}

To better understand how the choice of threshold $\tau$ influences the dynamic properties of the constructed temporal graphs, we analyze the edge evolution across time using the Temporal Edge Appearance (TEA) plot~\cite{poursafaei2022towards}. The TEA plot quantifies, at each time point, the proportion of newly appearing edges compared to the previous graph snapshot, thereby offering a visual measure of temporal variability. Fig.~\ref{fig:TEA} shows how the choice of threshold $\tau$ affects the temporal dynamics of functional connectivity graphs. When $\tau$ is too small, the resulting graphs are overly dense and smooth, capturing many weak or noisy connections. This leads to a large number of repeated edges across time and consequently a lower average novelty score. In contrast, when $\tau$ is too large, each graph contains very few edges, making the connectivity patterns highly sparse and more dynamic, but potentially less stable. Thresholds around 0.4 and 0.5 provide a good balance, capturing meaningful changes while avoiding excessive noise. We further validate this observation in our ablation study, which shows that performance is sensitive to $\tau$, and based on these results, we select $\tau = 0.5$ for our final model. 

\section{Evaluation Metrics}
\label{appendix:lp_metrics}

\noindent
{\bf (1) Mean Average Precision (MAP):} For each query node $q$, predictions are ranked in descending order by their predicted probabilities. The average precision ($AP$) is calculated using Eq.~\ref{eq:map}, where $P(k)$ is the precision at rank $k$, and $rel(k)$ is 1 if the $k$-th item is relevant. The final $MAP$ is the mean of $AP$ values across all nodes in $Q$.

\begin{equation}
\text{MAP} = \frac{1}{|Q|} \sum_{q \in Q} AP(q);\quad AP(q) = \frac{1}{m} \sum_{k=1}^{n} \left[ P(k) \times rel(k) \right]
\tag{1}
\label{eq:map}
\end{equation}

\noindent
{\bf (2) Mean Reciprocal Rank (MRR):} 
$MRR$ is the average of reciprocal ranks of the first relevant prediction for each query node $q$, as shown in Eq.~\ref{eq:mrr}. Unlike $MAP$, which considers all relevant items, $MRR$ focuses on the rank of the first correct prediction.

\begin{equation}
\text{MRR} = \frac{1}{|Q|} \sum_{q \in Q} \frac{1}{k_q}
\tag{2}
\label{eq:mrr}
\end{equation}

\noindent
For F1-score, predicted link probabilities are binarized using an optimal threshold selected on the validation set. Specifically, for each experimental setting, the threshold that maximizes the validation F1-score is chosen and then applied to the corresponding test set. AUC and PR-AUC are computed in a threshold-free manner based on ranked prediction scores.

\section{Implementation Details}
\label{appendix:details}

We use 5 NVIDIA A100 GPUs with 8GB of memory each. Our model is implemented using the mamba-ssm library~\cite{dao2024transformers}. The Mamba2 temporal module is configured with hidden dimension $d_\text{model} = 96$, state dimension $d_\text{state} = 64$, convolution kernel size $d_\text{conv} = 4$, and head dimension set to 16. Our models were trained for 50 epochs using the AdamW optimizer with a weight decay of $8 \times 10^{-5}$. The learning rate was adjusted based on the generated graph sequences and varied between $10^{-6}$ and $10^{-7}$ across different experimental settings. To regularize the model during training, a dropout rate of 0.38 was applied after the Mamba block. To prevent overfitting, we also employed early stopping with a patience of 5 epochs.

For the baseline model TransformerG2G, the input sequence of graph embeddings with adaptive lookback is first projected to a $d$-dimensional space using a linear layer, where we set $d = 256$. The model consists of 8 encoder layers, each with a single attention head. A $\tanh$ activation is applied in the nonlinear layer following the encoder, and the resulting representation is further projected to the $D_\text{emb}$-dimensional space, consistent with our BrainATCL architecture. In the nonlinear projection head, we apply an $\text{ELU}$ activation function. The TransformerG2G model is optimized using the Adam optimizer with a fixed learning rate of $10^{-6}$ across all datasets. For the $\mathcal{GDG}$-Mamba baseline, we adopt the same training setup as used for our proposed BrainATCL model. For all adaptive lookback settings, we set $\ell_{\min} = 1$ and $\ell_{\max} = 4$.

We also compared BrainATCL against several static and dynamic GNNs for link prediction and age estimation tasks. 
Static GNNs include GCN \cite{kipf2016semi}, GraphSAGE \cite{hamilton2017inductive}, GAT \cite{velivckovic2017graph}, and GIN \cite{xu2018powerful}. 
Dynamic GNNs comprise compositions of static GNNs with gated recurrent units, GCN-GRU, GraphSAGE-GRU, GAT-GRU, and GIN-GRU \cite{chen2022gc,seo2018structured}, as well as the variational graph recurrent neural network (VGRNN) \cite{hajiramezanali2019variational}.

Both static and dynamic GNNs were trained end-to-end for both tasks. 
For static GNNs, the input is a static graph $G^{p,q}_k$, and the target is either the next graph $G^{p,q}_{k+1}$ or the subject’s age $y^p$. 
For dynamic GNNs, the input is a graph sequence of fixed lookback size four, $(G^{p,q}_{k-4}, \dots, G^{p,q}_k)$, with the same targets.

For the \textit{temporal link prediction task}, we trained an MLP classifier consists of one hidden layer with ReLU activation, as defined in Section~\ref{sec:BrainATCL-Framework}, to predict whether a pair of nodes is connected based on their embeddings.
The classifier was trained to minimize the weighted binary cross-entropy loss using an Adam optimizer with a learning rate of $10^{-3}$ and evaluated using MAP and MRR. For this task, each subject’s data was split into train/validation/test sets along the temporal dimension, following the same split setting used for training the BrainATCL. The reported MAP and MRR scores are averaged over all test graphs to evaluate embedding quality. For F1-score, the decision threshold is selected based on validation performance. 

In the \textit{age estimation task}, we ensured that subjects with familial relationships (e.g., twins) were assigned to the same dataset partition (training or testing) to avoid data leakage. This guarantees that related individuals do not appear in both sets simultaneously during model training and evaluation. We use a separate set of 300 subjects that are not involved in the link estimation experiments to avoid information leakage. The data is split at the subject level, with 80\% of the subjects used for training and 20\% for testing. All available sessions are included for each subject. Results are averaged over five runs, each with a different random 80/20 train-test split. At first, we generate a temporal embedding for each subject by concatenating the graph embeddings from all sessions. Before feeding these subject-level temporal embeddings into the age regression model, we apply a dimensionality reduction module to extract compact subject-level representations. Given an input tensor of shape $(P, K, N, D_\text{emb})$, where $P$ is the number of subjects, $K$ is the number of temporal graphs per subject, $N$ is the number of brain regions, and $D_\text{emb}$ is the embedding dimension, we first perform mean pooling over the region dimension to obtain a tensor of shape $(P, K, D_\text{emb})$. The resulting sequence is then passed through a 1D convolutional network consisting of
two convolutional layers with ReLU activations and an adaptive average pooling layer to aggregate temporal information into a fixed-size vector. For the $K$-dimensional representation, since it combines four sessions, we also experimented with an alternative approach in which we first averaged the four sessions before applying the convolutional module. In addition, we tested removing the first 20\% of graphs from each session, as the initial segment often contains substantial noise and fluctuations. We selected the best-performing configuration as the final result. Finally, a fully connected layer maps the output to a 256-dimensional representation, which is then used for downstream age prediction.


\section{Ablation study on the effect of hyperparameters}
\label{appendix:Ablation_p}

\begin{figure}[ht]
  \centering
\includegraphics[width=0.9\textwidth]{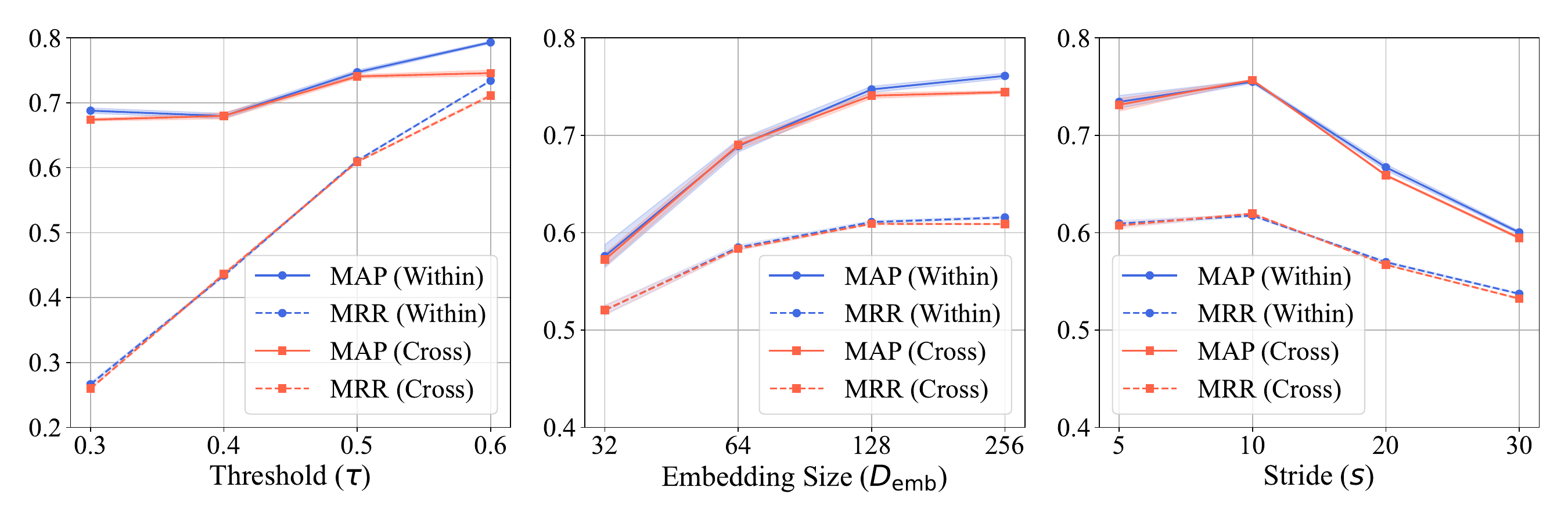}
  \begin{minipage}[htbp]{0.45\textwidth}
    \centering
    \includegraphics[width=\textwidth]{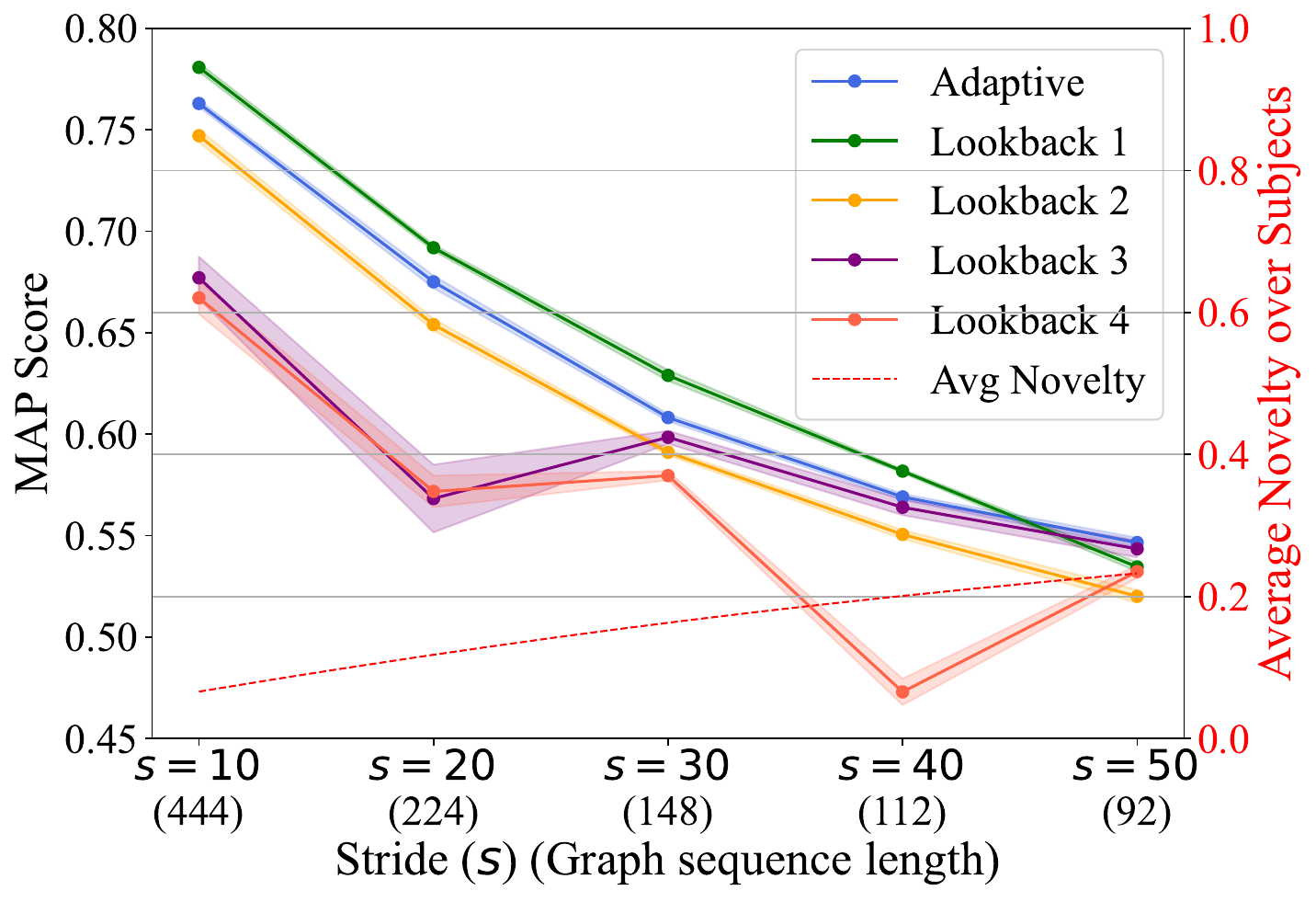}
  \end{minipage}
  \begin{minipage}[htbp]{0.45\textwidth}
    \centering
    \includegraphics[width=\textwidth]{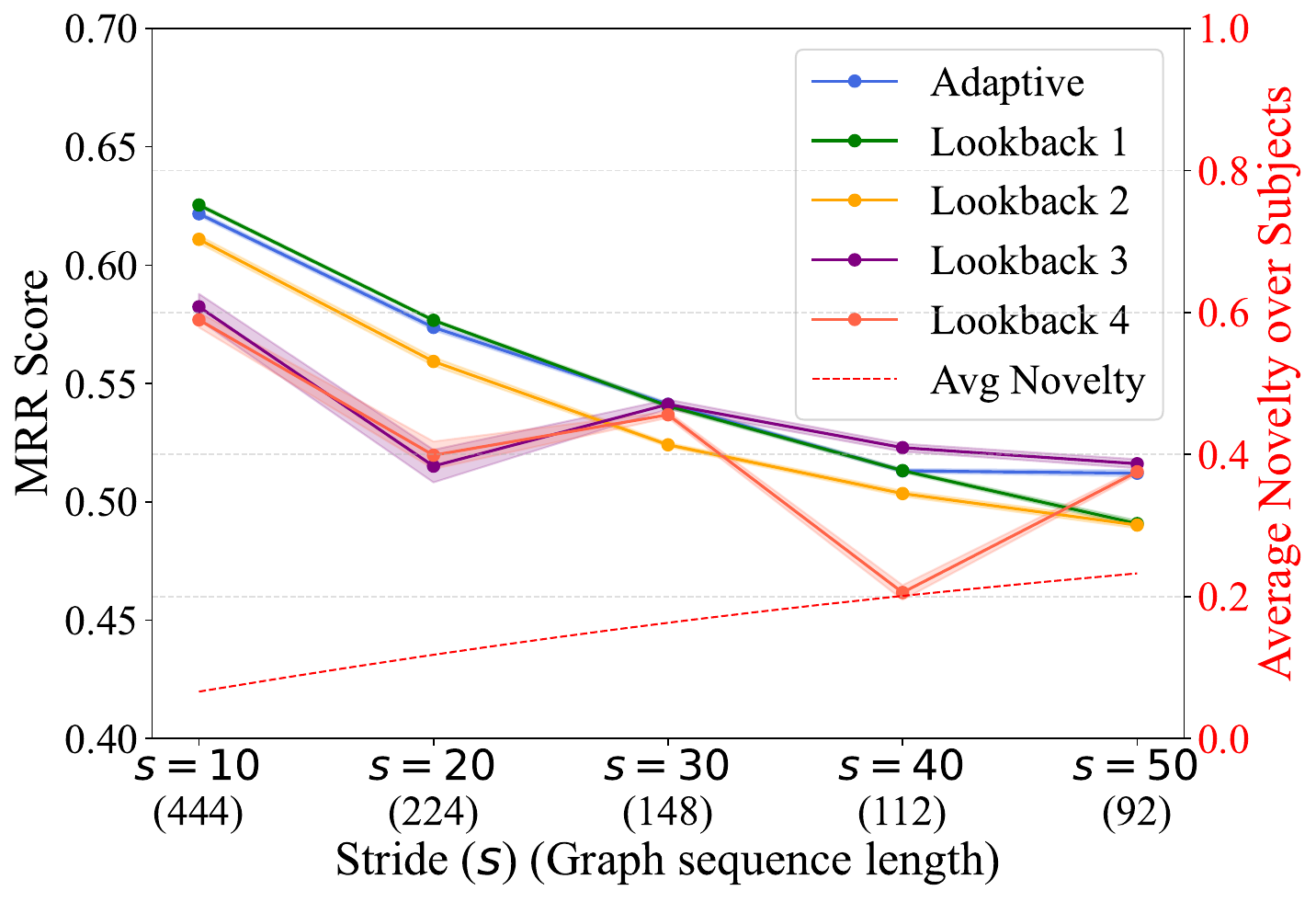}
  \end{minipage}

  \caption{\textbf{The Top panel} shows ablation study on the effect of hyperparameters (threshold, embedding size and stride) on \textit{within-session} (blue) and \textit{cross-session} (red) temporal link prediction performance in terms of MAP (solid lines) and MRR (dashed lines). Solid/dashed lines denote the mean performance, while the shaded areas correspond to the standard deviation across five runs. \textbf{The bottom panel} shows MAP and MRR scores across different lookback strategies and stride sizes for the link prediction task. The red dashed line shows the average novelty score corresponding to each stride setting. The x-axis denotes the stride size ($s$), with the number of temporal graphs per subject indicated in parentheses.}
  \label{fig:thred_embsize_stepsize}
\end{figure}

See Fig.~\ref{fig:thred_embsize_stepsize}.

\section{Explainability}
\label{appendix:explain}
\begin{figure}[htbp]
  \centering
  \includegraphics[width=1\textwidth]{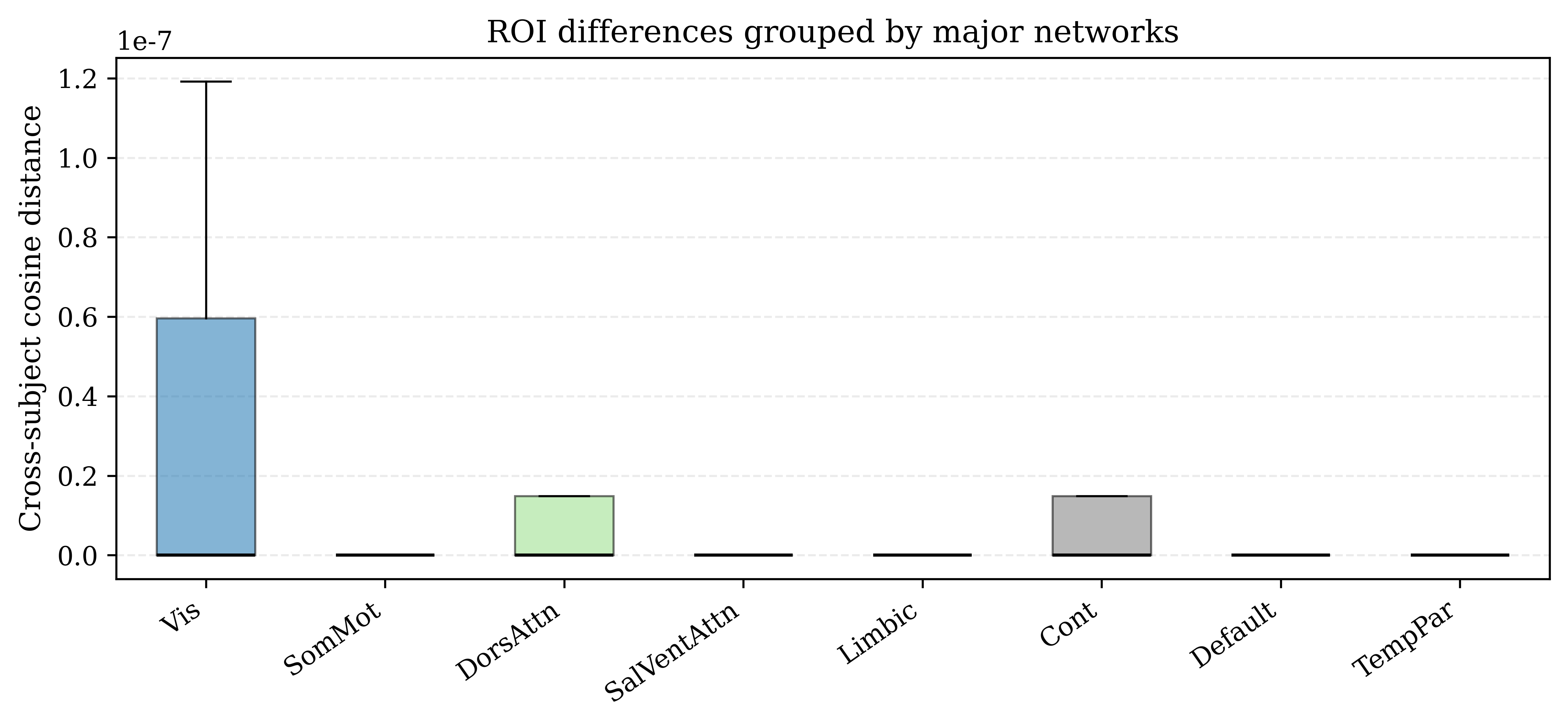}
  \caption{Distribution of cross-subject cosine distances between time-averaged ROI embeddings of the two subjects with the largest age difference (15 years), grouped by major functional networks. All subnetworks are defined based on the Schaefer parcellation~\cite{schaefer2018local}}

  \label{fig:explain}
\end{figure}

See Fig.~\ref{fig:explain}. Each box summarizes the distribution of ROI-level cosine distances within a functional network. This visualization reflects the heterogeneity of cross-subject embedding differences across ROIs within each network.

\end{document}